# DIAGNOSIS IS NOT A POLICY: A STATE-TRANSITION AUDIT OF CAPABILITY-CONDITIONED TEST-TIME INFERENCE

Yuchong Hu[1], Zhao Zhang[1], Kainan Liu[2], Haoyu Wang[1], Siyan Zhang[3], Jing Xiao[2,*]
[1] University of Science and Technology of China [2] Ping An Technology (Shenzhen) Co., Ltd., China [3] Western University
* Corresponding author: Jing Xiao
xiaojing661@pingan.com.cn

## ABSTRACT

Test-time actions can satisfy retrieval, state-realization, program-consistency, or rendering objectives without correcting an answer. Test-time Error Response and Realizability Audit (TERRA) is an evaluation/auditing protocol, not a new inference algorithm. It uses a unified decomposition of intervention outcomes into applicability, target-state realization, endpoint translation, harm, and resource/inference cost. Paired comparisons fix the incoming state and evaluation conditions; cross-model analyses vary the receiving configuration. Executable financial reasoning experiments yield three findings: the same context expansion can harm one configuration and benefit another; sufficient evidence does not ensure correctness because configurations differ in evidence-to-answer translation; and diagnostic headroom—accuracy attainable with idealized diagnostic information over the measured baseline—plus local consistency does not ensure practical gains. Action value is V(a|x,m), not diagnosis alone; diagnosis is not a deployable policy π(a|x,m).

## 1. INTRODUCTION

Test-time scaling allocates additional computation, information, or search during inference rather than shifting capability from trained parameters to inference alone. Prior work studies retrieval depth adapted to question complexity [1], context sufficiency [2], model routing [3], search and reasoning-compute allocation [4], [5], and capability regimes for verification and self-correction [6]–[9]. These studies establish that question difficulty, context sufficiency, and model capability affect action benefits. We focus on executable states before and after an action, separately measuring target-state realization and the receiving model's translation of that state into a correct task outcome.

Context expansion, targeted retrieval, retry, verification, and rendering are predefined interventions on an execution trace. Each targets a specified trace property; target-state realization and endpoint correctness are evaluated separately. Define $V_\lambda(a \mid x,m) = \mathbb{E}[Y^a - Y^0 - \lambda^T(\boldsymbol{C}^a - \boldsymbol{C}^0) \mid x,m]$, where x is the complete incoming state, a is the action, m is the receiving configuration, and λ contains resource weights. Hence action value depends on the incoming state, action mechanism, and receiving model. An action can be inapplicable, fail to realize its target, or realize a local objective without yielding a correct answer. TERRA asks whether an action is applicable, whether its target is realized, whether the model translates that state correctly, and what harm and resource cost arise; final accuracy collapses these outcomes.

Related work can be organized as intervention→diagnosis→correction. Intervention systems allocate retrieval depth, model routing, search, and inference compute [1]–[5]. Sufficient Context separates context sufficiency from failures to use it [2]. Other diagnosis studies identify error locations, verification dynamics, and conditions under which retrieval can repair financial reasoning [7], [8], [10]. Correction work examines self-correction limits, capability thresholds and overcorrection, and financial reasoning or verification pipelines [6], [9], [11]–[13]. Across these lines, action execution and endpoint correctness are usually summarized together; none systematically separates whether an intervention was applicable and realized from whether the receiving model translated the realized state into a correct endpoint. TERRA supplies that audit layer and additionally distinguishes oracle diagnostics from deployable action selection.

We audit FinQA/DocFinQA official/test and an independent train-300 mechanism set with six configurations spanning the Qwen and Gemma families. TERRA is an offline evaluation/auditing protocol, not an inference algorithm. It contributes a unified decomposition of applicability, realization, endpoint translation, harm, and resource cost; an oracle-versus-deployable diagnosis distinction; and paired evidence that context expansion has configuration-dependent effects, sufficient evidence does not ensure endpoint correctness, and diagnostic headroom plus local consistency does not ensure practical gains. Official/cross-family evaluation supports the first finding; mechanism and oracle-versus-practical comparisons support the others.

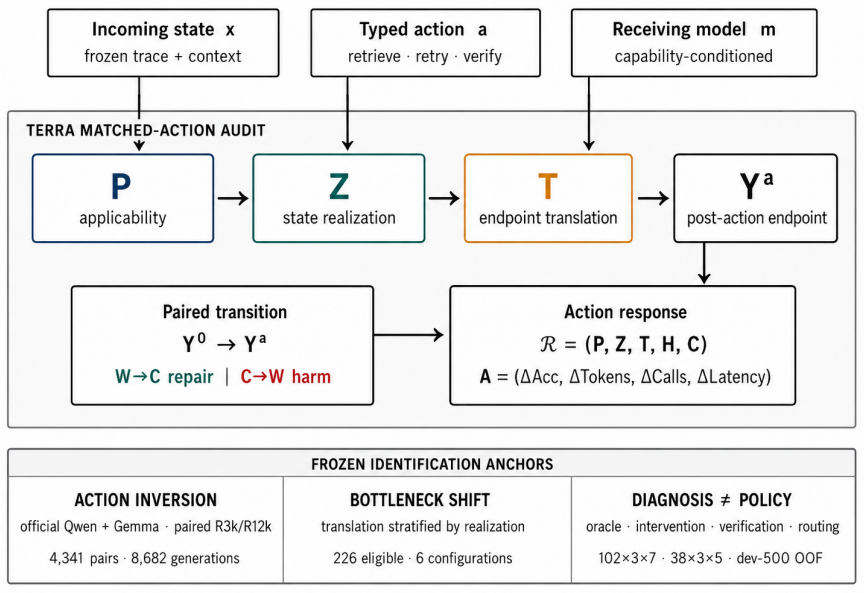

Fig. 1. TERRA audit protocol and experimental anchors.

## 2. TERRA: AN EVALUATION AND AUDITING PROTOCOL

### 2.1. Action Transitions, Diagnosis, and Policy

Let $x_i$ denote the pre-intervention state: question $q_i$, context, trace/program, instruction template, and endpoint inputs. The triplet $(x,a,m)$ specifies state, action, and receiving configuration. Within $m$, paired runs compare baseline $a_0$ with action $a$, holding $x$ and evaluation conditions fixed except for action-specified changes. Within each experiment, cross-configuration comparisons share questions and action specifications; generated traces may differ. Oracle diagnosis labels failure or intervention eligibility. An inference-time policy $\pi(a \mid x,m)$ selects an action, including no action; diagnosis alone does not specify a policy.

### 2.2. Response Decomposition and Net Effect

For question i, configuration m, and action a, let $Y^{0}_{i,m}$ and $Y^{a}_{i,m} \in \{0,1\}$ denote the pre- and post-action task outcomes. Define the action-response vector

$$\mathscr{R}_{i,a,m} = (P_{i,a}, Z_{i,a}, T_{i,a,m}, H_{i,a,m}, \mathbf{C}_{i,a,m}) \qquad (1)$$

$P \in \{0,1\}$ denotes membership in the action-specific audit cohort, not execution. Ungated actions may run outside this cohort; eligible-only experiments provide no outcomes for unexecuted cases. For audited transitions, $Z = 1$ iff the post-action trace meets the target criterion. $T = 1\{Y^{a} = 1\}$ is a sample-level endpoint-translation indicator defined only for $Z = 1$; otherwise $T$ is undefined. Its conditional mean is $\tau = \Pr(T = 1 \mid Z = 1, \mathscr{E}_{a})$. Harm is $H = 1\{Y^{0} = 1, Y^{a} = 0\}$. The vector $\boldsymbol{C} \in [0,\infty)^{3}$ records token, call, and latency resource proxies, not monetary price. For context expansion, Z requires all reference-program operands and their metric, period, entity, unit, and provenance roles; other actions use explicit realization criteria.

Let $\mathscr{E}_{a} = \{i : P_{i,a} = 1\}$ be the action-specific eligible cohort. All probabilities below use this same cohort: $b_{m,a} = \Pr(Y^{0} = 0 \mid \mathscr{E}_{a})$, $r_{a,m} = \Pr(Y^{a} = 1 \mid Y^{0} = 0, \mathscr{E}_{a})$, and $h_{a,m} = \Pr(Y^{a} = 0 \mid Y^{0} = 1, \mathscr{E}_{a})$. The net accuracy effect is

$$\Delta\mathrm{Acc}(a,m;\mathscr{E}_{a}) = b_{m,a} r_{a,m} - (1 - b_{m,a}) h_{a,m} \qquad (2)$$

Equation (2) weights repair gains and regression losses by the baseline proportions of incorrect and correct answers. Full-set effects apply (2) to all evaluated cases, including those outside $\mathscr{E}_{a}$ under ungated execution. We report accuracy and cost changes relative to the same paired baseline $a_{0}$:

$$\mathbf{A}(a,m;\mathscr{E}_{a}) = (\Delta\mathrm{Acc}, \Delta\mathrm{Tokens}, \Delta\mathrm{Calls}, \Delta\mathrm{Latency}) \qquad (3)$$

ΔAcc, ΔTokens, ΔCalls, and ΔLatency denote changes in accuracy and resource-cost proxies over $\mathscr{E}_{a}$; C changes are measured proxy differences, not monetary prices. Pareto comparisons require shared endpoints, cohorts, and proxy definitions. Configuration-dependent repair and harm rates can reverse the net effect. Repair rates within Z=1 and Z=0, with the Z=0 share of unrepaired errors, characterize residual bottlenecks.

## 3. EXECUTABLE FINANCIAL REASONING AND EXPERIMENTAL PROTOCOL

### 3.1. Data, Models, and Actions

FinQA supplies report questions, supporting facts, executable programs, and answers [14]; DocFinQA aligns questions with full reports [15]. Dev-500 has 500 questions in 188 report clusters; train-300 has 300 questions in 146 firm-year clusters, with no report-cluster overlap. Official/test has 1,147 questions (Table V).

Qwen3 8B/14B and Qwen3.8 27B support official-test and mechanism experiments; Gemma 3 4B/12B/27B provide cross-family replication. The 3,441 official-test and 1,800 mechanism pairs total 5,241 pairs and 10,482 generation cells. Each pair fixes question, model, prompt, and retrieval implementation, varying only the predefined action, once at temperature=0.

Rk denotes the retrieval-side evidence character budget. R3000 uses question-only lexical retrieval, top-k=8, and 3,000 characters. R12000 retrieves eight BM25 anchors, expands in anchor→previous→next order to top-k=24, and concatenates deterministically up to 12,000 characters. This generic context expansion changes both budget and adjacent-context retrieval. Other actions are targeted R9000, one empty-step retry, deterministic rendering, strict evidence updates, verify-before-act, and learned routing. Evidence, Structure, Reasoning, Certification, and Rendering denote the affected trace dimensions.

### 3.2. Labels, Endpoints, and Inference

The 226 R3000 evidence-deficient questions span 131 firm-year clusters. After R12000, 102 satisfy Section 2.2 (Z=1); the other 124 comprise 104 missing-evidence and 20 ambiguous-alignment cases (Z=0). Targeted retrieval addresses the 104 missing cases with all-operand rescue; empty retry targets 17 empty-step cases and answer–program consistency; rendering targets 37 executable cases with unequal original/rendered values. Z is assigned post hoc and Generic/Targeted eligibility uses gold labels: these are oracle audit strata, not runtime selectors. Deployable diagnosis must be predictable from x without gold or future information; ungated expansion, verify-before-act, and learned routing are deployment-facing probes.

We distinguish execution correctness YD, certified correctness YC, and displayed-answer correctness YR. Tables I, II, VII and the first three Table III actions use YD; rendering uses YR. Table I is full-set, Table II stratifies initially incorrect cases, and Table III reports explicit numerator/denominator counts for action-specific eligible cohorts. Bootstrap procedures are in Section 5.1.

**TABLE I**

Execution accuracy and context-expansion gains

| Config. | R3k→R12k | ΔAcc [95% CI] | W→C/C→W |
|---|---|---|---|
| Off. Q8B (1147) | 11.07→9.85 | −1.22 [−2.84,0.35] | 33/47 |
| Off. Q14B (1147) | 26.16→20.40 | −5.75 [−7.96,−3.56] | 49/115 |
| Off. Q27B (1147) | 60.68→62.51 | +1.83 [0.16,3.52] | 57/36 |
| Gemma 4B (300) | 3.67→1.00 | −2.67 [−4.62,−1.00] | 0/8 |
| Gemma 12B (300) | 5.00→6.00 | +1.00 [−1.72,3.83] | 11/8 |
| Gemma 27B (300) | 13.00→27.33 | +14.33 [9.48,19.27] | 52/9 |

Note: Official-test Qwen: n=1,147 per configuration; Gemma replication: n=300 per configuration.

## 4. RESULTS AND ANALYSIS

### 4.1. The Same Action Can Harm or Help Different Configurations

Table I reports paired official-test Qwen and Gemma replication. Qwen 14B and Gemma 4B are negative, both 27B models are positive, and Qwen 8B/Gemma 12B intervals cross zero. The Qwen 27B−14B contrast is +7.59 pp (synchronized bootstrap CI [5.00,10.22]); Gemma 27B−4B is +17.00 pp ([12.08,22.26]).

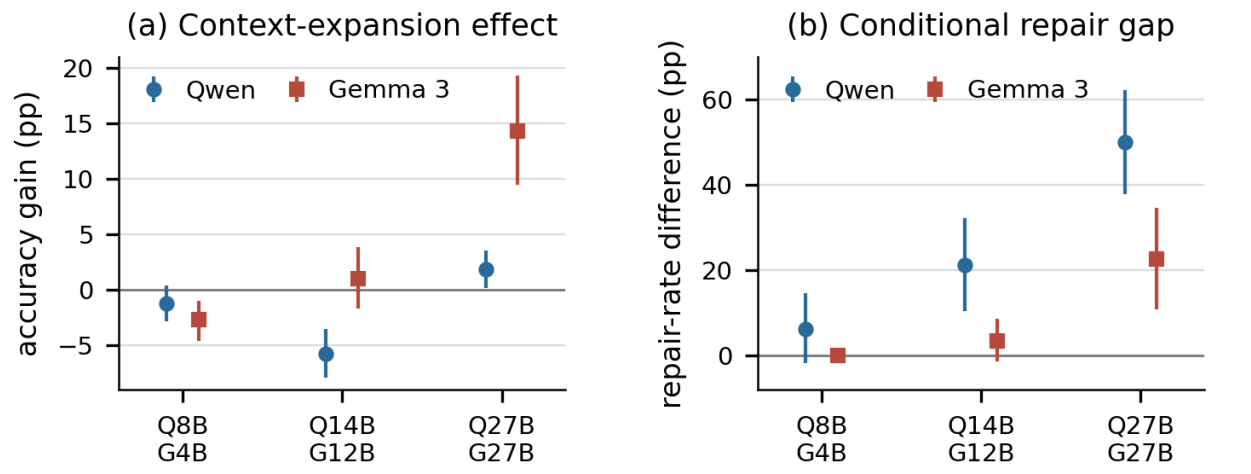

Fig. 2. Context-expansion effects (left) and repair-rate gaps (right), with 95% CIs.

### 4.2. Sufficient Evidence Does Not Ensure a Correct Answer

**TABLE II**

Repair rates and residual errors by evidence stratum

| Config. | Repair $r_1/r_0$ K/N (%) | Repair gap $r_1-r_0$ (pp) | Z=0 among residual errors n/N (%) |
|---|---|---|---|
| Qwen 8B | 12/101 (11.88%)<br>7/122 (5.74%) | 6.14 | 115/204 (56.37%) |
| Qwen 14B | 32/100 (32.00%)<br>13/121 (10.74%) | 21.26 | 108/176 (61.36%) |
| Qwen 27B | 66/95 (69.47%)<br>23/118 (19.49%) | 49.98 | 95/124 (76.61%) |
| Gemma 4B | 0/102 (0.00%)<br>0/124 (0.00%) | 0 | 124/226 (54.87%) |
| Gemma 12B | 5/100 (5.00%)<br>2/123 (1.63%) | 3.37 | 121/216 (56.02%) |
| Gemma 27B | 33/99 (33.33%)<br>13/121 (10.74%) | 22.59 | 108/174 (62.07%) |

In Table II, upper/lower K/N entries denote Z=1/Z=0 among initially incorrect cases. Define $r(z) = \Pr(Y^a = 1 \mid Y^0 = 0, Z = z, \mathscr{E}_a)$, $R_z = N_z - K_z$, and the Z=0 residual share $R_0/(R_0 + R_1)$. Qwen 27B's Z=1 denominator 95 excludes seven initially correct cases; both Gemma 4B strata are nonempty. These conditional rates differ from Table III's Repair/Realized column, which counts baseline-incorrect W→C within the realized subset. Despite realized evidence sufficiency, 29/95 (30.53%) initially incorrect Qwen 27B cases and 66/99 (66.67%) Gemma 27B cases remain wrong. For Qwen 27B, 95/124 (76.61%) residual errors occur in $Z = 0$.

### 4.3. Diagnostic Headroom Does Not Ensure Practical Correction

Local state success does not guarantee endpoint gains: one dev-500 empty-step retry makes 44/54 traces answer–program consistent but only 1/54 task-correct. The 17 Table III cases are from independent train-300 and are not pooled with these 54. Structural validity and local consistency are insufficient for semantic correctness.

Table IV separates oracle headroom from practical interventions. The largest oracle jump is evidence→binding, but practical evidence-fixed actions do not match it. Identifying a key state does not ensure that an action realizes it or that the receiving model translates it correctly.

**TABLE III**

Action outcomes with explicit denominators

| Action base/end | Eligible | Realized | W→C | C→W | Net/ Eligible | Repair/ Realized |
|---|---|---|---|---|---|---|
| Generic R12k R3k / YD | 226/300 | 102/226 | 89/226 | 2/226 | +87/226 | 66/102 |
| Targeted R9k R12k / YD | 104/300 | 9/104 | 5/104 | 2/104 | +3/104 | 4/9 |
| Empty retry R3k / YD | 17/300 | 12/17 | 1/17 | 0/17 | +1/17 | 1/12 |
| Det. render R12k / YR | 37/300 | 37/37 | 9/37 | 2/37 | +7/37 | 9/37 |

Note: Every count is numerator/denominator. Eligible=cohort/300; Realized/Eligible; W→C and C→W/Eligible. Repair/Realized counts baseline-incorrect W→C among realized cases, excluding cases that remain correct: Generic has 89 eligible W→C, only 66 among its 102 realized cases. Net/Eligible=(W→C−C→W)/Eligible; differing baselines/endpoints preclude cross-action ranking.

All actions use Qwen3.8-27B-FP8. Table III's Generic row is the 226-question gold-deficient cohort: 89 W→C, 2 C→W, +87/226. The full set has 93 W→C and 7 C→W (+86/300=+28.67 pp in YD); the 74 noneligible cases contribute 4/5. The counts differ by cohort scope, not contradictory measurement. Targeted/Empty are eligible-only. Ungated rendering yields 9/3 over 300 (+2.00 pp), versus 9/2 gated; the extra regression maps a non-renderable case to empty.

Rendering changes textual answers, not $Y_D$. For STT/2007, the displayed value changes from 108288.89 to 10828888.89; for RSG/2013, 225.0% becomes 22500.0%. Both programs execute but return incorrect results: $Y_D$ remains incorrect while $Y_R$ changes from correct to incorrect. Zero model calls do not imply zero harm to the displayed-answer endpoint.

**TABLE IV**

Oracle headroom versus practical interventions

| Evidence type | Design | Key result |
|---|---|---|
| Oracle evidence→ binding | 102×3×7 | +43.14/+42.16/+17.65pp |
| Evidence-fixed | 102×3 | +4.90/−1.96/+1.96pp |
| Strict CF Evidence Update | 38×3×5 | Original 2/11/24 → updated 2/2/2 (out of 38) |
| Verify-before-Act | 102×3×3 | Weak/mid negative; strong ≈+0.98 pp |
| Learned routing | dev-500 OOF | Reachability predictable; no Pareto gain |

## 5. FROZEN CONFIRMATION, COST, AND SCOPE

### 5.1. Frozen Evaluation and Statistical Protocol

Supplementary analyses use 10,000 paired cluster-bootstrap replicates (PCG64, lexicographic order, linear quantiles), with cluster effect 100Σsg/Σng. Official-test intervals use 380 report-ID groups; firm-year regrouping gives a Qwen-contrast CI of [5.18,10.03] pp. Fig. 2(right), on the train-300 mechanism set, samples all 146 firm-year clusters before eligibility/stratum filtering with common draws. These post hoc intervals supplement the frozen Fig. 2(left) CIs; Gemma 4B's [0,0] is degenerate because it has zero repairs.

**TABLE V**

Frozen official-test protocol (intention-to-treat)

| Item | Frozen setting |
|---|---|
| Official split | 1,147 questions; 380 report-ID file groups; 278 firm-year clusters; 2,294 contexts |
| Generation cells | 6,882 scored cells; 15 generation failures retained as incorrect under ITT |
| Decoding | temperature=0; max tokens=192; enable thinking=false |
| Identification | Paired cluster bootstrap; 10,000 draws; original seed 20260825 |
| Evaluation | Single evaluation; frozen samples, prompts, revisions, and scoring |

All 6,882 official-test cells are retained under intention-to-treat: generation failure is the primary outcome and is coded incorrect, not imputed. This includes five R12000 failures per configuration. Requests, responses, scores, mappings, and manifests form the reproducibility record.

### 5.2. Cost and Conditional Transitions

TABLE VI

Context-expansion resource-cost proxies on dev-500 (Qwen3.8-27B-FP8)

| Resource proxy | R3000 | R12000 | Change |
|---|---|---|---|
| Mean prompt tokens/request | 2,308 | 5,060 | 2.19× |
| Mean serving latency | 3.25 s | 4.38 s | +34.8% |
| Total model tokens, 500 questions | 1.197 M | 2.581 M | 2.16× |
| Retrieval replay latency/question | 27.1 ms | 38.7 ms | +42.8% |

R12000 increases input length, serving latency, and token usage (Table VI). These quantities are resource-cost proxies, not estimates of monetary cost; accuracy effects vary across receiving configurations.

TABLE VII

Conditional repair and harm rates on official/test

| Config. | Repair r n/N (%) | Harm h n/N (%) | G=r−h (pp) |
|---|---|---|---|
| 8B | 33/1020 (3.24%) | 47/127 (37.01%) | −33.77 |
| 14B | 49/847 (5.79%) | 115/300 (38.33%) | −32.55 |
| 27B | 57/451 (12.64%) | 36/696 (5.17%) | +7.47 |

Table VII reports repair r among initially incorrect cases and harm h among initially correct cases. G=r−h is a conditional-rate difference; ΔAcc additionally requires the baseline weights in (2).

### 5.3. Evidence Binding and Evaluator Validation

Program leaves are classified as document values, valid constants, or intermediate variables, then numerically matched within gold supporting facts. Certified binding checks value, metric, period, entity, unit, and provenance; duplicate sources form equivalence sets, and ambiguous alignments are excluded. Out-of-set evidence IDs are removed and logged without changing program or gold fields. The frozen evaluator incorporates schema-recovery and valid-constant fixes (Table VIII(a)). Fig. 3 distinguishes individual diagnostic coverage from joint pass rates.

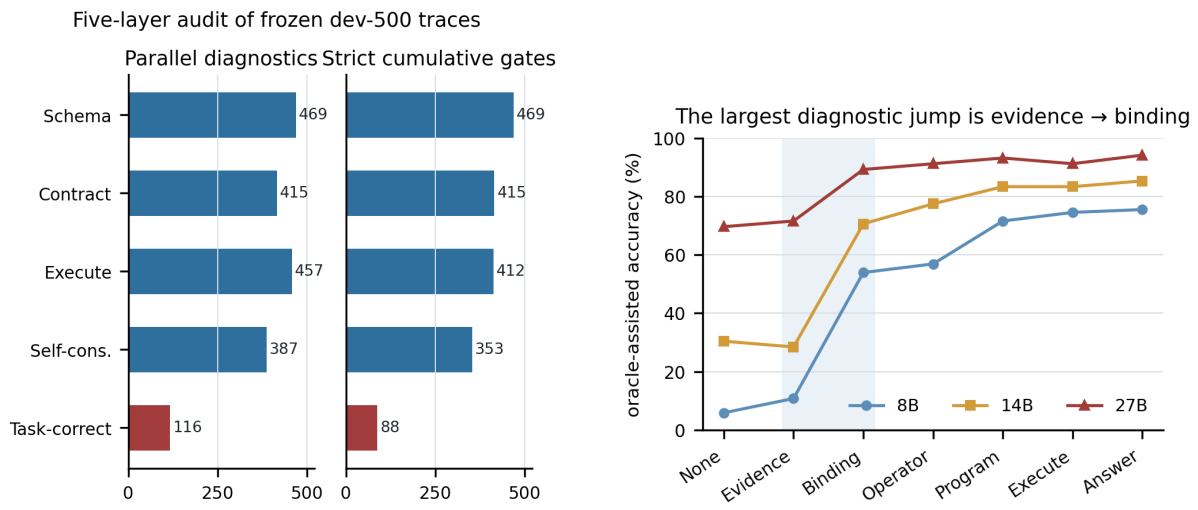

Fig. 3. Diagnostic coverage (left) and oracle ladder (right).

The oracle ladder isolates the largest gain at evidence→binding (Fig. 3); all stages assume idealized intermediate information.

### 5.4. Evidence Hierarchy and Scope of Claims

Table VIII groups supporting evidence and reports a post hoc comparison standardized to the development-set evidence composition.

TABLE VIII

Supporting evidence and evidence-composition standardization

**(a) Frozen confirmation and mechanism evidence**

| Experiment | Evidence |
|---|---|
| Context expansion | Official Qwen (n=1,147): harm at 8B/14B, benefit at 27B. Gemma (n=300): harm at 4B, benefit at 27B. |
| Mechanism experiments | On train-300, configurations differ in repair rates after evidence becomes sufficient. |
| Four action families | State realization, correction, and harm are measured separately (Table III). |
| Oracle ladder | On 102 questions, evidence-to-operand binding gives the largest oracle gain. |
| Evaluator validation | Schema recovery: 86.4→87.6%; valid-constant handling: 73.6→79.0%. |

**(b) Stress tests and deployment boundaries**

| Probe/control | Finding |
|---|---|
| Strict update and verification | Realizing a target state rarely yields a correct answer (Table IV). |
| Learned routing | dev-500: reachability AUC 0.649/AP 0.705; routing yields no Pareto improvement. |
| Budget, order, and conflict | Effects depend on intervention settings. |
| Repair granularity | 469 traces: local repairability does not imply whole-trace repairability. |
| Rendering control | dev-500 correct answers: 70→84 (R3000), 142→159 (R12000); train-300 still shows harm. |

**(c) Evidence-composition standardization (Qwen 27B)**

| Dataset | Exec. accuracy (%) | Raw gain (pp) | Initially deficient | Deficient→sufficient | Std. gain (pp) |
|---|---|---|---|---|---|
| dev-500 | 23.20→46.00 | +22.80 | 378/500 (75.60%) | 140/378 (37.04%) | +22.80 |
| train-300 | 21.33→50.00 | +28.67 | 226/300 (75.33%) | 102/226 (45.13%) | +28.77 |

Table VIII(c) reanalyzes R3000→R12000 without new model calls. Realization rates use initially deficient cases as the denominator. Standardizing stratum-specific gains to the dev-500 composition (75.60% deficient, 24.40% other) leaves a 5.97 pp gap, unexplained by this composition. Independent contributions of baseline accuracy and realization remain unidentified. These post hoc estimates exclude official/test.

The evidence supports descriptive comparisons across the tested configurations: context expansion changes both budget and retrieval, and the Qwen checkpoints span generations. Offi-cial/test combines 929 uniquely matched full-report contexts with 218 page-level fallbacks.

## 6. CONCLUSION

Official-test and cross-family results show that context expansion can shift from net harm to net benefit across receiving configurations; mechanism experiments show configuration-dependent translation and post-realization bottlenecks. TERRA is an evaluation/auditing protocol, not a new inference method. It asks whether an action is applicable, realizes its target, translates correctly, and has acceptable harm/resource cost. The findings suggest distinguishing evidence acquisition from utilization when allocating computation. Diagnosis locates repairs, but oracle headroom and local consistency do not establish deployable gains; deployment requires measured net endpoint gains and component-wise reporting. The protocol is domain-agnostic by design; current evidence is financial-reasoning-only.